\documentclass[11pt,a4paper]{article}
\usepackage[hyperref]{acl2019}
\usepackage{times}
\usepackage{latexsym}

\usepackage{url}

\usepackage[utf8]{inputenc}
\usepackage[english]{babel}

\aclfinalcopy 
\def\aclpaperid{1274} 

\title{SherLIiC: A Typed Event-Focused Lexical Inference Benchmark
  for Evaluating Natural Language Inference}

\author{Martin Schmitt \and Hinrich Schütze \\
  Center for Information and Language Processing (CIS)\\
  LMU Munich, Germany\\
  {\tt martin@cis.lmu.de}
}

\date{}

\usepackage{graphicx}
\usepackage{amssymb}
\usepackage{amsmath}
\usepackage{amsthm}
\usepackage{dsfont}
\usepackage[nameinlink,capitalise]{cleveref}
\usepackage{textcomp}
\usepackage{microtype}
\usepackage{booktabs}
\usepackage{tabularx}
\newcolumntype{Y}{>{\centering\arraybackslash}X}

\usepackage{multirow}
\usepackage[inline]{enumitem}

\newcolumntype{R}{>{$}r<{$}} 
\newcolumntype{C}{>{$}c<{$}} 
\newcolumntype{L}{>{$}l<{$}} 

\usepackage{dcolumn}

\DeclareMathVersion{nxbold}
\SetSymbolFont{operators}{nxbold}{OT1}{cmr} {b}{n}
\SetSymbolFont{letters}  {nxbold}{OML}{cmm} {b}{it}
\SetSymbolFont{symbols}  {nxbold}{OMS}{cmsy}{b}{n}

\newcolumntype{.}{D{.}{.}{3}}
\makeatletter
\newcolumntype{B}{>{\mathversion{nxbold}\DC@{.}{.}{3}}c<{\DC@end}}
\makeatother

\newcommand{\newcol}[2]{\multicolumn{1}{#1}{#2}}

\crefname{section}{§}{§§}
\Crefname{section}{§}{§§}

\newcommand{\R}{\mathcal{R}}

\newcommand{\E}{\mathcal{E}}
\newcommand{\T}{\mathcal{T}}

\newcommand{\abs}[1]{\left\lvert#1\right\rvert}

\newcommand{\textapprox}{\texttildelow}

\newcommand{\minrelsize}{r_\text{min}}

\newcommand{\blfont}[1]{\texttt{#1}}

\newcommand{\minip}[1]{\noindent\textbf{#1}}

\newcommand{\eset}[1]{\left\{ #1 \right\}}
\newcommand{\cset}[2]{\left\{\, #1 \mid #2 \,\right\}}
\newcommand{\pair}[1]{\left( #1 \right)}
\DeclareMathOperator*{\argmax}{arg\,max}

\graphicspath{{./img/}}

\def\resource{SherLIiC}

\def\candidates{\resource{}-InfCands}
\def\testset{\resource{}-test}
\def\devset{\resource{}-dev}
\def\relext{\resource{}-TEG}

\usepackage{tikz}

\makeatletter
\newcommand*{\radiobutton}{%
  \begin{tikzpicture}
    \pgfmathsetlengthmacro\radius{height("X")/2.5}
    \draw[radius=\radius,
      preaction={
        draw=gray,
        transform canvas={
          xshift=.7\pgflinewidth,
          yshift=-.7\pgflinewidth,
        },
      },  
      preaction={fill=white},
    ] circle;
  \end{tikzpicture}%
}
\makeatother

\def\dnrm#1{\mbox{$_{\hbox{\scriptsize #1}}$}}

\begin{document}
\maketitle

\begin{abstract}

  We present \resource{},\footnote{\url{https://github.com/mnschmit/SherLIiC}}
  a testbed for lexical inference in context (LIiC),
  consisting of
  3985  manually annotated \emph{inference rule candidates} (InfCands),
  accompanied by 
  (i) \texttildelow{}960k unlabeled InfCands,
  and
  (ii) \texttildelow{}190k \emph{typed textual relations} between Freebase entities
  extracted from the large entity-linked corpus ClueWeb09.
  Each InfCand consists of one of these relations,
  expressed as a lemmatized dependency path,
  and two argument placeholders,
  each linked to one or more Freebase types.
  Due to our candidate selection process based on strong distributional evidence,
  \resource{} is much harder than existing testbeds
  because distributional evidence is of little utility
  in the classification of InfCands.
  We also show that, due to its construction,
  many of \resource{}'s correct InfCands are novel and
  missing from existing rule bases.
  We evaluate a number of strong baselines on \resource{},
  ranging from semantic vector space models
  to state of the art neural models of natural language inference (NLI).
  We show that \resource{} poses a tough
  challenge to existing NLI systems.

\end{abstract}


\section{Introduction}
\label{sec:introduction}

Lexical inference (LI) can be seen as a focused variant
of natural language inference (NLI), also called 
recognizing textual entailment \citep{dagan13}.
Recently, \citet{gururangan18} showed that
annotation artifacts in current NLI testbeds distort
our impression of the performance of state of the art systems,
giving rise to the need for new evaluation methods for NLI.
\citet{glockner18} investigated LI as a way of evaluating NLI systems and
found that even simple cases are challenging to
current systems.
In this paper, we release \resource{}, a testbed
specifically designed
for evaluating a system's
ability to solve the hard problem of
modeling lexical entailment
in context.

\begin{table}
  \setlength\tabcolsep{0.25em}
  \def\arraystretch{1.2}
  \small
  \centering
  \begin{tabularx}{\linewidth}{ll|X}
    \multirow{2}{*}{(1)} & \multirow{2}{*}{troponymy} &\textsc{orgf}[A] is granting to \textsc{empl}[B]\\
         && $\Rightarrow$ \textsc{orgf}[A] is giving to \textsc{empl}[B]\\
    \hline
        \multirow{2}{*}{(2)} & synonymy + & \textsc{orgf}[A] is supporter of \textsc{orgf}[B] \\
        & derivation & $\Rightarrow$ \textsc{orgf}[A] is backing \textsc{orgf}[B] \\\hline
        \multirow{2}{*}{(3)} & typical & \textsc{auth}[A] is president of \textsc{loc}[B]\\
        & actions & $\Rightarrow$ \textsc{auth}[A] is representing \textsc{loc}[B]\\\hline
        \multirow{2}{*}{(4)} & script & \textsc{per}[A] is interviewing \textsc{auth}[B] \\
        & knowledge & $\Rightarrow$ \textsc{per}[A] is asking \textsc{auth}[B] \\\hline
    \multirow{2}{*}{(5)} & common sense &\textsc{orgf}[A] claims \textsc{loc}[B] \\
         &knowledge& $\Rightarrow$ \textsc{orgf}[A] is wanting \textsc{loc}[B]\\
  \end{tabularx}
  \caption{Examples of \resource{} InfCands and NLI challenges they cover. \textsc{orgf}=organization founder, \textsc{empl}=employer, \textsc{auth}=book author, \textsc{loc}=location, \textsc{pol}=politician, \textsc{per}=person.}
  \label{tab:phenomena}
\end{table}

\citet{levy16} identified context-sensitive
-- as opposed to ``context-free'' --
entailment
as an important evaluation criterion
and created a dataset for LI in context (LIiC).
In their data, WordNet \citep{wordnet95,wordnet05} synsets serve as context
for one side of a binary relation,
but the other side is still instantiated with a single concrete expression.
We aim to improve this setting in two ways.

First, we type our relations on
\emph{both sides},
thus making them more general.
Types provide a context
that can help in disambiguation
and at the same time allow generalization over contexts because
arguments of the same type are represented abstractly.
An example for the need for disambiguation is
the verb ``{run}''. ``{run}'' entails 
``{lead}'' in the context of \textsc{person} /
\textsc{company} (``Bezos runs Amazon''). But in the context of \textsc{computer}
/ \textsc{software}, ``{run}'' entails ``{execute}''/``{use}''  (``my mac runs mac\-OS'').  Here, types help find the
right interpretation.

Second, \emph{we only consider relations between named entities} (NEs).
Inference mining based on non-NE types such as
WordNet synsets
(e.g.,  \textsc{animal}, \textsc{plant life})
primarily discovers \emph{facts} like
``parrotfish feed on algae''.
In contrast, 
the focus on NEs makes it more
likely that we will capture \emph{events} like ``Walmart closes gap
with Amazon''
and thus knowledge about event entailment
like [``$A$ is closing gap with $B$''
$\Rightarrow$ ``{$B$ is having lead over $A$}''] that 
is substantially
different from knowledge about general facts.

In more detail, we create \resource{} as follows.
First,
we extract verbal relations between Freebase \citep{freebase}
entities from the entity-linked web corpus ClueWeb09
\citep{gabrilovich13}.\footnote{\url{http://lemurproject.org/clueweb09}}
We then divide these relations into typable subrelations
based on the most frequent Freebase types found in their
extensions.
We then create
a large set of
inference rule candidates (InfCands), i.e.,
premise-hypothesis-pairs of verbally
expressed relations.
Finally, 
we use Amazon Mechanical Turk to classify
each InfCand in a randomly sampled subset
as \emph{entailment} or \emph{non-entailment}.

In summary, our contributions are the following:
\begin{enumerate*}[label={(\arabic*)}]
\item We create
\resource, a new resource for LIiC, consisting of
3985 {manually annotated InfCands}.
Additionally,
we provide
\texttildelow{}960k {unlabeled InfCands} ({\candidates{}}),
and
the {typed event graph} {\relext},
containing \texttildelow{}190k typed textual binary relations between Freebase entities.
\item \resource{}
is {harder} than existing testbeds because
{distributional evidence} is of {limited utility} 
in the classification of InfCands.
Thus, \resource{} is a promising and
challenging resource for developing NLI systems that go
beyond shallow semantics.
\item Human-interpretable knowledge graph types
serve as context for both sides of InfCands.
This makes InfCands more general
and
boosts the number of event-like relations in \resource{}.
\item \resource{} is {complementary to existing collections} of inference rules
as evidenced by the low recall these resources achieve (cf.\ \cref{tab:performance}).
\item We evaluate a large number of baselines on
\resource.
The best-performing baseline makes use of typing.
\item We demonstrate that
{existing NLI systems do poorly} on
\resource.
\end{enumerate*}


\section{Generation of InfCands}
\label{sec:data-generation}

This section  describes
 creation (\cref{sec:relation-extraction})
and typing (\cref{sec:typing})
of the typed event graph \relext{}
and then the generation of \candidates{} (\cref{sec:entailment-discovery}).

\subsection{Relation Extraction}
\label{sec:relation-extraction}

For each sentence $s$ in ClueWeb09 that contains at least two entity mentions,
we use MaltParser \citep{malt07} to generate a dependency graph,
where nodes are labeled with their lemmas and edges with dependency types.
We take all shortest paths between all combinations of two entities in $s$
and represent them by alternating edge and node labels.
As we want to focus on relations that express events,
we only keep paths with a nominal subject on one end.
We also apply heuristics to filter out erroneous parses.
See \cref{app:relation-filter} for  heuristics and
\cref{tab:metarules} for examples of relations.

\minip{Notation.}
Let $\R$ denote the set of extracted relations.
A relation $R \in \R$ is represented as a set of pairs of Freebase entities (its extension): $R \subseteq \E \times \E$, with $\E$ the set of Freebase entities.
Let $\pi_1, \pi_2$ be functions that map a pair to its first or second entry, respectively.
By abuse of notation, we also apply them to sets of pairs.
Finally, let $\T$ be the set of Freebase types and $\tau \colon \E \to 2^{\T}$ the function that maps an entity to the set of its types.

\subsection{Typing}
\label{sec:typing}

We define a \emph{typable subrelation} of  $R \in \R$
as a subrelation
whose entities in each argument slot share at least one type, i.e., an $S \subseteq R$ such that:
\begin{displaymath}
  \forall i \in \eset{1, 2} \colon \exists t \in \T \colon t \in \bigcap_{e \in \pi_i(S)} \tau(e)
\end{displaymath}
We compute the set $\operatorname{Type}_{k^2}(R)$ of
the (up to) $k^2$ largest typable subrelations of $R$ and
use them instead of $R$.
First, for each argument slot $i$
of the binary relation $R$,
the $k$ types $t^i_j$ (with $1 \leq j \leq k$) are computed that occur most often in this slot:
\begin{displaymath}
  \displaystyle t^i_j := \argmax_t \abs{\cset{p\in R}{t \in \tau_j^i(\pi_i(p))}}
\end{displaymath}
with
\begin{align*}
  \tau_1^i(e) &= \tau(e)\\
  \tau_{j+1}^i(e) &= \tau_j^i(e) - \eset{t^i_j}
\end{align*}
Then, for each pair
\begin{displaymath}
  \pair{s, u} \in \cset{\pair{t^1_j, t^2_l}}{j, l \in \eset{1, \dots, k}}
\end{displaymath}
of these types, we construct a subrelation
\begin{displaymath}
  R_{s,u} := \cset{\pair{e_1, e_2}\in R}{s\in\tau(e_1), u\in\tau(e_2)}
\end{displaymath}
If $\abs{R_{s,u}} \geq \minrelsize$, $R_{s,u}$ is included in $\operatorname{Type}_{k^2}(R)$.
In our experiments, we set $k = \minrelsize = 5$.

The type signature ({tsg}) of a typed relation $T$ is
defined as the pair of sets of types that is common to
first (resp.\ second) entities in the extension:
\begin{displaymath}
  \operatorname{tsg}(T) = \pair{\bigcap_{e\in\pi_1(T)} \tau(e), \bigcap_{e\in\pi_2(T)} \tau(e)}  
\end{displaymath}

\minip{Incomplete type information.}
Like all large knowledge bases, Freebase suffers from incompleteness:
Many entities have no type.
To avoid losing information about relations associated with
such entities,
we introduce a special type $\top$ and define $\argmax_t \abs{\emptyset} := \top$.
We define the relations $R_{s, \top}$, $R_{\top, u}$ and
$R_{\top, \top}$
to have no type restriction on entities in a $\top$ slot.
This concerns approximately 17.6\%{} of the relations in \relext{}.

\subsection{Entailment Discovery}
\label{sec:entailment-discovery}
Our discovery procedure is based on
Sherlock \citep{schoenmackers10}.
For the InfCand $A \Rightarrow B$ ($A, B \in \R$),
we define the relevance score $\operatorname{Relv}$,
a metric expressing Sherlock's stat.\ relevance criterion $P(B \mid A) \gg P(B)$ \citep[cf.][]{salmon71}.
\begin{displaymath}
\operatorname{Relv}(A, B) := \frac{P(B \mid A)}{P(B)} = \frac{\abs{A\cap B} \abs{\E\times\E}}{\abs{A} \abs{B}}
\end{displaymath}
Our significance score  $\sigma(A, B)$
is a scaled version of the significance test $\operatorname{lrs}$ used by Sherlock:
\begin{displaymath}
  P(B \mid A) \operatorname{lrs}(A, B) = \frac{\abs{A\cap B}\operatorname{lrs}(A, B)}{\abs{A}}
\end{displaymath}
with $\operatorname{lrs}(A, B)$ (likelihood ratio statistic) defined as
\begin{displaymath}
  2 \abs{A} \sum_{H\in\eset{B, \neg B}} P(H \mid A)  \log(\operatorname{Relv}(A, H)).
\end{displaymath}
Additionally, we introduce the \emph{entity support ratio}:
\begin{displaymath}
  \operatorname{esr}(A, B) := \frac{\abs{\bigcup_{i\in\eset{1,2}} \pi_i(A\cap B)}}{2\abs{A\cap B}}
\end{displaymath}
This score measures the diversity of
entities in $A\cap B$.
We found that
many InfCands involve a few frequent entities and
so obtain high $\operatorname{Relv}$ and $\sigma$ scores even though
the relations of the rule are semantically unrelated.
$\operatorname{esr}$
penalizes such InfCands.

We apply our three scores
defined above
to all possible pairs of relations
$(A, B) \in \R\times\R$ and accept a rule
iff all of the following criteria are met:
\begin{enumerate}[topsep=7pt,itemsep=2.5pt,parsep=2.5pt]
\item $\forall i \in \eset{1,2} \colon \pi_i(\operatorname{tsg}(A\Rightarrow B)) \neq \emptyset$
\item $|A \cap B| \geq \minrelsize$
\item $\forall i \in \eset{1,2} \colon  |\pi_i(A \cap B)| \geq \minrelsize$
\item $\operatorname{Relv}(A, B) \geq \vartheta_\textit{relv}$
\item $\sigma(A, B) \geq \vartheta_\sigma$
\item $\operatorname{esr}(A, B) \geq \vartheta\dnrm{esr}$
\end{enumerate}
where $\operatorname{tsg}(A\Rightarrow B)$ is  component-wise intersection of $\operatorname{tsg}(A)$ and $\operatorname{tsg}(B)$
and $\vartheta_\textit{relv}=1000$, $\vartheta_{\sigma}=15$, $\vartheta\dnrm{esr}=0.6$.
We found these numbers by randomly sampling InfCands
and inspecting their scores.
Typing lets us set these thresholds  higher,
benefitting
the quality of \candidates{}.

Lastly, we apply  \citet{schoenmackers10}'s heuristic to
only accept the 100 best-scoring premises for each hypothesis.
For each hypothesis $B$, we rank all possible premises $A$ by the product of the three scores
and filter out cases where $A$ and $B$ only differ in their types.


\def\crowdspace{0.5cm}
\begin{figure}
  \fontsize{8.7pt}{10.5pt}\selectfont
  \textbf{Fact}: location[B] is annexing location[A]\,.\\
  \mbox{\phantom{aa}}Examples for location[B]: \textit{Russia / USA / Indonesia} \\
  \mbox{\phantom{aa}}Examples for location[A]: \textit{Cuba / Algeria / Crimea} \\[0.2cm]
  $\square$ fact incomprehensible\\[0.2cm]
  \textbf{Please answer the following questions:}\\
  \begin{tabularx}{\linewidth}{|X|}
    \hline Is it {certain} that location[B] is taking control of location[A]?\\
    \hspace{\crowdspace} \radiobutton{} yes \hspace{\crowdspace} \radiobutton{} no  \hspace{\crowdspace} \radiobutton{} incomprehensible\\\hline
    Is it {certain} that location[B] is taking 
    location[A]?\\
    \hspace{\crowdspace} \radiobutton{} yes \hspace{\crowdspace} \radiobutton{} no  \hspace{\crowdspace} \radiobutton{} incomprehensible\\\hline
    Is it {certain} that location[B] is bordered by
    location[A]?\\
    \hspace{\crowdspace} \radiobutton{} yes \hspace{\crowdspace} \radiobutton{} no \hspace{\crowdspace} \radiobutton{} incomprehensible\\
    \hline
  \end{tabularx}
  \caption{Annotation Interface on Amazon MTurk}
  \label{fig:hit}
\end{figure}

\section{Crowdsourced Annotation}
\label{sec:annotation}

\candidates{} contains \textapprox{}960k InfCands.
We take a random sample of size 5267
and annotate it using  Amazon Mechanical Turk (MTurk).

\subsection{Task Formulation}
\label{sec:task-formulation}
We are asking our workers the same kind of questions
as \citet{levy16} did.
We likewise form batches of sentence pairs
to reduce annotation cost.
Instead of framing the task as judging the appropriateness of answers,
however,
we state the premise as a fact and ask workers
about its entailed consequences, i.e., we ask
for each candidate hypothesis
whether it is \emph{certain} that it is also true.
\cref{fig:hit} shows the annotation interface schematically.

We use a morphological lexicon  \citep{xtag01}
and form the present tense of a dependency path's predicate.
If a sentence is flagged incomprehensible (e.g., due to a parse error),
it is excluded from further evaluation.

We put  premise and hypothesis in the present (progressive
if suitable) based on
the intuition that a pair is only to be considered an entailment
if the premise makes it necessary for the hypothesis to be
true \emph{at the same time}.
This condition of simultaneity
follows the tradition of datasets such as SNLI
\citep{snli} -- in contrast to more lenient
evaluation schemes that consider a rule to be correct
if the hypothesis is true at any time before or after the
reference time of the premise
\citep[cf.][]{lin01,lewis13}.

\begin{table}
  \small
  \centering
  \begin{tabularx}{1.0\linewidth}{Xr}
    \toprule
    \multicolumn{2}{l}{\textbf{Annotated Subset of \candidates{}}}\\
    Validated InfCands & 3985\\
    Balance yes/no & 33\%{} / 67\%{}\\
    Pairs with unanimous gold label & 53.0\%{}\\
    Pairs with 1 disagreeing annotation & 27.4\%{}\\
    Pairs with 2 disagreeing annotations & 19.6\%{}\\
    Individual label = gold label & 86.7\%{}\\
    \bottomrule
  \end{tabularx}
  \caption{Statistics for crowd-annotated InfCands. The gold label is the majority label.}
  \label{tab:annotation_stats}
\end{table}

\subsection{Annotation Quality}
\label{sec:annot-qual}
We imposed several qualification criteria on crowd workers:
number of previous jobs on MTurk,  overall acceptance rate and
a  test that each worker had to pass.
Some workers still had frequent low agreement with the
majority.
However, in most cases we obtained a clear majority annotation.
These annotations were then used to automatically detect workers with low \emph{trust},
where \emph{trust} is defined as the ratio of submissions agreeing with the majority answer and a worker's total number of submissions.
We excluded workers with a \emph{trust} of less than $0.8$ and
collected replacement annotations until
we had at least five annotations per InfCand.

\begin{figure}
  \centering
  \includegraphics[width=\linewidth]{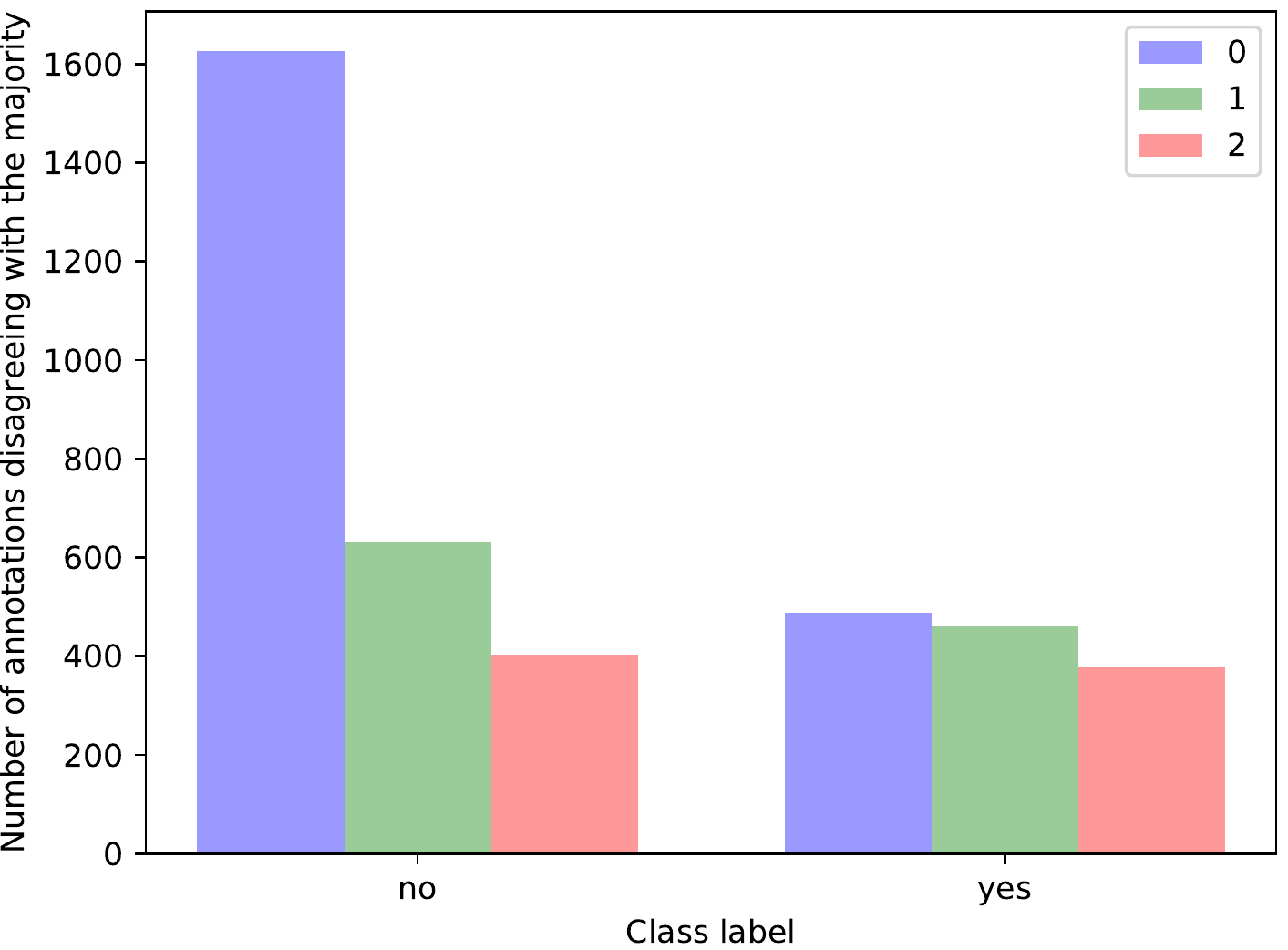}
  \caption{Number of disagreements per class label on all annotations.}
  \label{fig:disagr_per_class}
\end{figure}

\cref{tab:annotation_stats} shows that 
workers agreed unanimously for 53\% and that the maximal
number of two disagreements only occurs for 19.6\%.
The high number of times an individual agrees with the gold label suggests that humans can do the task reliably.
Interestingly, the number of disagreements is not evenly distributed among the two classes \emph{entailment}/\emph{non-entailment}.
If the majority agrees on \emph{entailment},
it is comparatively much more likely that at least one of the workers disagrees (cf.\ \cref{fig:disagr_per_class}).
This suggests that our workers were  strict and keen on achieving  high precision in their annotations.


\section{Baselines}
\label{sec:baselines}

We split our annotated data 25:75 into \devset{} and \testset{},
stratifying  on  annotated label (\emph{yes}/\emph{no}) and number of disagreements with the majority (0/1/2).
If unanimity of annotations marks particularly clear cases,
we figure they should be evenly distributed on dev and test.

To establish the state of the art for \resource{},
we evaluate a number of baselines.
Input to these baselines are either the dependency paths
or the sentences that were presented to the crowd workers.
Baselines that require a threshold to be used as binary classifiers are tuned on \devset{}.

\minip{Lemma baseline.}
Following \citet{levy16}, this baseline classifies an InfCand as valid
if the following holds true for the premise $p$ and hypothesis $h$ after lemmatization:
\begin{enumerate*}[label={(\arabic*)}]
\item $p$ contains all of $h$'s content words,\footnote{We use the stop word list of \texttt{nltk} \citep{nltk}.}
\item $p$'s and $h$'s predicates are identical, and
\item the relations' active/passive voice matches their arguments' alignment.
\end{enumerate*}

\minip{Rule collection baselines.}
\blfont{Berant I} \citep{berant11} and \blfont{Berant II} \citep{berantThesis}\footnote{We use default threshold 0.}
are entailment graphs.
\blfont{PPDB} is the largest collection (XXXL) of PPDB 2.0 \citep{ppdb}.\footnote{We ignore stop words and punctuation for phrases.}
\blfont{Patty} is a collection of relational patterns,
consisting of ontological types, POS placeholders and words.
We use the version extracted from Wikipedia with Freebase types \citep{patty}.
\blfont{Schoenmackers} is the rule collection released by \citet{schoenmackers10}.
\blfont{Chirps} is an ever-growing\footnote{We use the version downloaded on May 28, 2019.} predicate paraphrase database
extracted
via event coreference in news Tweets \citep{shwartz17-coref}. 
\blfont{All Rules} denotes the union of all of these rule bases.
For rules with type (or POS) constraints,
we ignore these constraints to boost recall.
We will see that even with these recall-enhancing measures,
the majority of our correct InfCands is not covered by existing rule bases.

\minip{Word2vec baselines.}
\blfont{word2vec} is based on \citet{word2vec_emb}'s
pre-trained word embeddings of size 300.
We average them to obtain a vector representation of relations consisting of multiple words
and use cosine similarity to judge a relation pair.
\blfont{typed\_{}rel\_{}emb} (resp.\ \blfont{untyped\_{}rel\_{}emb}) is obtained
by training word2vec skip-gram \citep{word2vec_alg}
with vector size 300 and otherwise default parameters
on a synthetic corpus representing
the extensions of typed (resp.\ untyped) relations.
The corpus is obtained by writing out one entity-relation-entity-triple per line,
where each entity is prefixed with the argument slot it belongs to.
\blfont{w2v+typed\_{}rel}
(resp.\ \blfont{w2v+untyped\_{}rel}) produces its score by
summing the scores of \blfont{word2vec} and
\blfont{typed\_{}rel\_{}emb} (resp.\ \blfont{untyped\_{}rel\_{}emb}).

Some type signatures ({tsg}s) benefit more from type-informed methods than others.
For example, the correct inference
[\emph{\textsc{influencer} is explaining in \textsc{written\_{}work}} $\Rightarrow$ \emph{\textsc{influencer} is writing in \textsc{written\_{}work}}]
is  detected by
\blfont{w2v+typed\_{}rel}, but not by \blfont{w2v+untyped\_{}rel}.
We therefore combine these two methods by using, for each
tsg, the method that works better for that tsg on dev.
(For {tsg}s not occurring in dev, we
take the method that works better
for the individual types occurring in the tsg.
We
use untyped embeddings if all else fails.)
We refer to this combination as \blfont{w2v+tsg\_{}rel\_{}emb}.

\minip{Knowledge graph embedding baselines.}
As \relext{} has the structure of a knowledge graph (KG),
we also evaluate the two KG embedding methods \blfont{TransE} \citep{transE}
and \blfont{ComplEx} \citep{complex},
as provided by the OpenKE framework \citep{openKE}.

\minip{Asymmetric baselines.}
Entailment models built upon cosine similarity are symmetric
whereas entailment is not.
Therefore many asymmmetric measures
based on the distributional inclusion hypothesis
have been proposed \citep{kotlerman10,santus14,shwartz17,roller18}.
We consider \blfont{Weeds\-Prec} \citep{weeds04} and \blfont{invCL} \citep{lenci12},
which have strong empirical results on hypernym detection.
We use cooccurrence counts with entity pairs as distributional representation of a relation.

\minip{Supervised NLI models.}
As LIiC is a special case of NLI, 
our dataset can also be used to evaluate
the generalization capabilities of supervised models trained on large NLI datasets.
We pick \blfont{ESIM} \citep{chen17}, a state-of-the-art supervised NLI model, trained on MultiNLI \citep{multinli} as provided by the framework Jack the Reader \citep{weissenborn18}.
Input to \blfont{ESIM} are the sentences from the annotation process
with placeholders instantiated by entities randomly picked from the example lists
that had also been shown to the crowd workers (cf.\ \cref{fig:hit}).
As we want to measure \blfont{ESIM}'s capacity to detect entailment,
we map its prediction of both
\emph{neutral} and \emph{contradiction} to our
\emph{non-entailment} class.

\minip{Sherlock+ESR.}
We also evaluate the candidate scoring method inspired by \citet{schoenmackers10}
that created the data in the first place.
We again combine the three scores
described in \cref{sec:data-generation}
by multiplication.
The low performance of \blfont{Sherlock+ESR} (cf.\ \cref{tab:performance})
is evidence that
the dataset is not strongly biased in its favor
and thus is promising as a general evaluation benchmark.


\begin{table*}[ht]
  \def\biggersep{6pt}
  \small
  \centering
  \begin{tabular}{rlp{\biggersep}.@{\,}Lp{\biggersep}...p{\biggersep}...}
    \toprule
    \multicolumn{6}{c}{} & \multicolumn{3}{c}{dev} && \multicolumn{3}{c}{test}\\
    \cmidrule(lr){7-9}\cmidrule(lr){11-13}
    & Baseline && \theta^* &&& \newcol{c}{P} & \newcol{c}{R} & \newcol{c}{F1} && \newcol{c}{P} & \newcol{c}{R} & \newcol{c}{F1}\\
    \midrule
    1&    \blfont{Berant I} && \text{--} &&&0.699&0.154&0.252&& 0.762 & 0.126  & 0.216 \\
    2&    \blfont{Berant II} &&\text{--}&&&0.800&0.181&0.296&& 0.774 & 0.186 & 0.300 \\
    3&    \blfont{PPDB} &&\text{--}&&&0.631&0.211&0.317&& 0.621 & 0.240 & 0.347 \\
    4&   \blfont{Patty} &&\text{--}&&&0.795&0.187&0.303&& 0.779 & 0.153 & 0.256 \\
    5&    \blfont{Schoenmackers} &&\text{--}&&& 0.780 & 0.139 & 0.236 && 0.849 & 0.119 & 0.208 \\
    6& \blfont{Chirps} && \text{--}&&& 0.370 & 0.308 & 0.336 && 0.341 & 0.295 & 0.316 \\
    7&    \blfont{All Rules} &&\text{--}&&&0.418&0.483&0.448&& 0.404 & 0.493 & 0.444 \\
    \midrule
    8&    \blfont{Lemma} &&\text{--}&&&\newcol{B}{0.900}&0.109&0.194&&\newcol{B}{0.907}& 0.089 & 0.161 \\
    9&    \blfont{Always yes} &&\text{--}&&&0.332&\newcol{B}{1.000}&0.499&& 0.333 &\newcol{B}{1.000}& 0.499 \\
    10&    \blfont{ESIM} &&\text{--}&&&0.391&0.831&0.532&& 0.390 & 0.833 & 0.531 \\
    \midrule
    11&    \blfont{word2vec} &&0.321&&&0.556&0.625&0.589&&0.520&0.606&0.559\\
    12&    \blfont{typed\_{}rel\_{}emb} &&0.864&&&0.561&0.568&0.565&& 0.532 & 0.486 & 0.508\\
    13&    \blfont{untyped\_{}rel\_{}emb} &&0.613&&&0.511&0.740&0.605&& 0.499 & 0.672 & 0.572\\
    \midrule
    14&    \blfont{w2v+typed\_{}rel} &&1.106&&&0.549&0.710&0.619&& 0.523 & 0.688 & 0.594\\
    15&    \blfont{w2v+untyped\_{}rel} &&0.884&&&0.565&0.740&0.641&& 0.528 & 0.695 & 0.600\\
    16&   \blfont{w2v+tsg\_{}rel\_{}emb} &&0.884&&&0.566&0.776&\newcol{B}{0.655}&& 0.518 & 0.727 &\newcol{B}{0.605}\\
    \midrule
    17&    \blfont{WeedsPrec (typed)} &&0.073&&&0.335&0.994&0.501&& 0.333 & 0.988 & 0.498\\
    18&    \blfont{WeedsPrec (untyped)} &&0.057&&&0.403&0.807&0.538&& 0.386 & 0.783 & 0.517\\
    19&    \blfont{invCL (typed)} &&0.000&&&0.332&\newcol{B}{1.000}&0.499&& 0.333 &\newcol{B}{1.000}& 0.499\\
    20&    \blfont{invCL (untyped)} &&0.148&&&0.362&0.876&0.512&& 0.357 & 0.863 & 0.505\\
    \midrule
    21&    \blfont{TransE (typed)} &&-0.922&&&0.336&\newcol{B}{1.000}&0.503&& 0.333 & 0.991 & 0.498\\
    22&    \blfont{TransE (untyped)} &&-0.476&&&0.340&0.964&0.503&& 0.332 & 0.942 & 0.491\\
    23&    \blfont{ComplEx (typed)} &&-0.033&&&0.339&0.955&0.500&& 0.337 & 0.949 & 0.497\\
    24&    \blfont{ComplEx (untyped)} &&-0.030&&&0.340&0.952&0.501&& 0.334 & 0.939 & 0.493\\
    \midrule
    25&    \blfont{Sherlock+ESR} &&9.460&{}\cdot 10^5&&0.504&0.592&0.544&& 0.491 & 0.526 & 0.508\\
    \bottomrule
  \end{tabular}
  \caption{Precision, recall and F1 score on \resource{}-dev and -test. All baselines run on top of \blfont{Lemma}. Thresholds ($\theta^*$) are F1-optimized on dev. Best result per column is set in bold.}
  \label{tab:performance}
\end{table*}

\section{Experimental Results and Discussion}
\label{sec:discussion}

\minip{Quantitative observations.}
\cref{tab:performance} summarizes the performance of our baselines
on predicting the \emph{entailment} class for \resource{}-dev and -test.

Rule collections (lines 1--6)
have recall between 0.119 and 0.308;
the recall of their union (line 7)
is only 0.483 on dev and 0.493 on test,
showing that we found indeed new valid inferences missing from existing rule bases.

The state-of-the-art neural model \blfont{ESIM} does not
generalize
well from MultiNLI (its training set) to LIiC.
In fact, it hardly improves on the baseline
that always predicts \emph{entailment} (\blfont{Always yes}).
Our dataset was specifically designed to only contain good InfCands based on distributional features.
So it poses a challenge to models that cannot make the fine semantic distinctions necessary for LIiC.

Turning to  vector space models (lines 11--24),
dense relation representations
(lines 12, 13)
predict entailment better than sparse models
(lines 17--20)
although they cannot use asymmetric measures.

KG embeddings
(lines 21--24)
do not seem at all appropriate for measuring the similarity of relations.
First, their performance is very close to \blfont{Always yes}.
Second, their F1-optimal thresholds are very low -- even negative.
This suggests that
their relation vectors do not contain any helpful information for the task.
These methods were not  developed to compare relations;
the lack of useful information is still surprising
and thus is a promising direction for future work on KG embeddings.

General purpose dense representations
(\blfont{word2vec}, line 11)
perform comparatively well,
showing that, in principle,
they cover  the information necessary
for LIiC.
Embeddings trained on our relation extensions \relext{}
(line 13), however,
can already alone achieve better performance than
word2vec embeddings alone.

In general, type-informed relation embeddings
seem to have a disadvantage compared to unrestricted ones
(e.g., cf.\ lines 12 and 13)
-- presumably because
type-informed baselines have training sets that are smaller
(due to filtering) and sparser (since relations are
split up according to type signatures). 
The combination of  general word2vec and specialized
relation embeddings
(lines 14--16),
however, consistently brings  gains.
This indicates that distributional word properties are
complementary to the relation extensions our method extracts.
So using both sources of information is promising
for future research on modeling relational semantics.

\blfont{w2v+tsg\_{}rel\_{}emb}
is the best-performing method.
It combines
typed and
untyped relation embeddings as well as general-purpose
word2vec embeddings.
Even though one cannot
rely on typed extensions only, this shows that incorporating
type information is beneficial for good performance.

We use \blfont{w2v+tsg\_{}rel\_{}emb}
to provide a noisy annotation for \candidates{}.
This is a useful resource because 
learning from noisy labels has been well studied
\citep{frenay14,hendrycks18} and is often beneficial.

\begin{table}
  \small
  \centering
  \begin{tabularx}{1.0\linewidth}{rX}
    \toprule
    \multirow{2}{*}{(1)} & \textsc{person}[A] is \textsc{region}[B]'s ruler \\
               & $\Rightarrow$ \textsc{person}[A] is dictator of \textsc{region}[B]\\
        \midrule
    \multirow{2}{*}{(2)} & \textsc{location}[A] is fighting with \textsc{orgf}[B] \\
    & $\Rightarrow$ \textsc{location}[A] is allied with \textsc{orgf}[B]\\
    \midrule
    \multirow{2}{*}{(3)} & \textsc{orgf}[A] is coming into \textsc{location}[B] \\
               & $\Rightarrow$ \textsc{orgf}[A] is remaining in \textsc{location}[B]\\
        \midrule
    \multirow{2}{*}{(4)} & \textsc{orgf}[A] is seeking from \textsc{orgf}[B] \\
    & $\Rightarrow$ \textsc{orgf}[B] is giving to \textsc{orgf}[A]\\
    \midrule
    \multirow{2}{*}{(5)} & \textsc{location}[A] is winning war against \textsc{location}[B] \\
    & $\Rightarrow$ \textsc{location}[A] is declaring war on \textsc{location}[B]\\
    \bottomrule
  \end{tabularx}
  \caption{False positives for each of the three best-performing baselines taken from \devset{}. \textsc{orgf}=organization founder.}
  \label{tab:false_positives}
\end{table}

\minip{Qualitative observations.}
Although \resource{}'s creation is based on the same method
that was used to create \blfont{Schoenmackers},
\resource{} is fundamentally different for several reasons:
\begin{enumerate*}[label={(\arabic*)}]
\item The rule sets are  complementary
  (cf.\ the low recall of $0.139$ and $0.119$ in \cref{tab:performance}).
\item The majority of rules in \blfont{Schoenmackers}
has more than one premise,
leaving only \textapprox{}13k InfCands in \blfont{Schoenmackers}
compared to \textapprox{}960k in  \candidates{}
that fit the format of NLI.
\item \blfont{Schoenmackers} is filtered more aggressively
with the goal of maximizing the number of correct rules.
This, however, makes it inadequate as a challenging benchmark
because the performance of \blfont{Always yes} would be close to 100\%.
\item \resource{} is focused on events.
When linking the relations from \relext{} back to their surface forms in the corpus,
80\%{} of them occur at least once in the progressive,
which suggests that the large majority of our relations indeed represent events.
\end{enumerate*}

Taking a closer look at \resource{},
we see that the data require a large variety of lexical knowledge
even though their creation has been entirely automatic.
\cref{tab:phenomena} shows five positively labeled examples from \devset{},
each highlighting a different challenge for statistical models
that is crucial for NLI.
(1) is an instance of  \emph{troponymy}:
``granting'' is a manner or kind of ``giving''.
This is the verbal equivalent to nominal hyponymy.
(2) combines synonymy (``support'' $\Leftrightarrow$
``back'') with morph.\ derivation.
(3) can only be classified correctly if one knows
that it is one of the typical actions of a president
to represent their country.
(4) requires knowledge about the typical course of events
when interviewing someone.
A typical interview involves asking questions.
(5) can only be detected
with common sense knowledge
that goes even beyond that:
you generally only claim something
if you want it.

An error analysis of the three best-performing baselines (lines 14--16 in \cref{tab:performance})
reveals that none of them was able to detect the five correct InfCands from \cref{tab:phenomena}.
Explicit modeling of one of the phenomena described above
seems a promising direction for future research to improve recall.
\cref{tab:false_positives} shows five cases
where InfCands were incorrectly labeled as \emph{entailment}.
(1) shows the importance of modeling directionality:
every ``dictator'' is a ``ruler'' but not vice versa.
(2) shows a well-known problem in representation learning from cooccurrence:
antonyms tend to be close in the embedding space \citep{mohammad08,mrkvsic16}.
The other examples show other types of correlation
that models relying entirely on distributional information will fall for:
the outcome of events like ``coming into a country'' or ``seeking something from someone'' are in general uncertain
although possible outcomes like
``remaining in said country'' (3) or ``being given the object of desire'' (4)
will be highly correlated with them.
Finally,
better models will also take into account the simultaneity constraint:
``winning a war'' and ``declaring a war'' (5)
rarely happen at the same time.


\def\metaspace{.3em}
\def\metacol{.7em}

\begin{table}
  \small
  \centering \begin{tabular}{@{}p{.59\linewidth}@{\hspace{\metacol}}c@{\hspace{\metacol}}p{.31\linewidth}@{}}
nsubj--\texttt{X}--prep--of--obj &$\Leftrightarrow$&  nsubj--\texttt{X}--poss\\
\emph{A is an \textbf{ally} of B} &&  \emph{A is B's \textbf{ally}}\\[\metaspace]
nsubj--\texttt{X}--prep--in--obj &$\Leftrightarrow$&  nsubj--\texttt{X}--poss\\
\emph{A is the \textbf{capital} in B} &&  \emph{A is B's \textbf{capital}}\\[\metaspace]
 nsubjpass--\texttt{X}--prep--by--obj &$\Leftrightarrow$& obj--\texttt{X}--nsubj\\
 \emph{A is \textbf{followed} by B} && \emph{B \textbf{follows} A}\\[\metaspace]
 nsubj--one--prep--of--obj--\texttt{X}--obj &$\Leftrightarrow$& nsubj--\texttt{X}--obj\\
 \emph{A is one of the \textbf{countries in} B} && \emph{A is a \textbf{country in} B}\\[\metaspace]
 nsubj--capital--conj--\texttt{X}--obj &$\Rightarrow$& nsubj--\texttt{X}--obj\\
 \emph{A is the capital and biggest \textbf{city in} B} && \emph{A is a \textbf{city in} B}\\\midrule
 nsubj--\texttt{X}er--prep--of--obj &$\Leftrightarrow$& nsubj--\texttt{X}--obj\\
 \emph{A is a \textbf{teach}er of B} && A \textbf{teaches} B\\[\metaspace]
 nsubj--co-\texttt{X}er--prep--of--obj &$\Rightarrow$& nsubj--\texttt{X}--obj\\
 \emph{A is a co-\textbf{found}er of B} && \emph{A \textbf{founds} B}\\[\metaspace]
 nsubj--re\texttt{X}--obj &$\Rightarrow$& nsubj--\texttt{X}--obj\\
 \emph{A re\textbf{writes} B} && \emph{A \textbf{writes} B}\\[\metaspace]
 nsubj--over\texttt{X}--obj &$\Rightarrow$& nsubj--\texttt{X}--obj\\
 \emph{A over\textbf{takes} B} && \emph{A \textbf{takes} B}\\\midrule
 nsubj--agree--xcomp--\texttt{X}--obj &$\Rightarrow$& nsubj--\texttt{X}--obj \\
 \emph{A agrees to \textbf{buy} B} && \emph{A \textbf{buys} B} \\[\metaspace]
 nsubjpass--force--xcomp--\texttt{X}--obj &$\Rightarrow$& nsubj--\texttt{X}--obj \\
 \emph{A is forced to \textbf{leave} B} && \emph{A \textbf{leaves} B}\\[\metaspace]
 nsubjpass--elect--xcomp--\texttt{X}--obj &$\Leftrightarrow$& nsubj--\texttt{X}--obj \\
 \emph{A is elected to be \textbf{governor} of B} && \emph{A is \textbf{governor} of B} \\[\metaspace]
 nsubj--go--xcomp--\texttt{X}--obj &$\Rightarrow$& nsubj--\texttt{X}--obj \\
 \emph{A is going to \textbf{beat} B} && \emph{A \textbf{beats} B}\\[\metaspace]
 nsubj--try--xcomp--\texttt{X}--obj &$\Rightarrow$& nsubj--\texttt{X}--obj \\
 \emph{A tries to \textbf{compete} with B} && \emph{A \textbf{competes} with B}\\[\metaspace]
 nsubj--decide--xcomp--\texttt{X}--obj &$\Rightarrow$& nsubj--\texttt{X}--obj \\
 \emph{A decides to \textbf{move to} B} && \emph{A \textbf{moves to} B} \\[\metaspace]
 nsubjpass--expect--xcomp--\texttt{X}--obj &$\Rightarrow$& nsubj--\texttt{X}--obj \\
 \emph{A is expected to \textbf{visit} B} && \emph{A \textbf{visits} B}
  \end{tabular}
  \caption{Most frequent meta rules (top), character level
    meta rules (middle), and implicative verb meta rules
    (bottom). Bold: Words corresponding to  \texttt{X}.}
  \label{tab:metarules}
\end{table}

\section{Meta Rules and Implicative Verbs}
\label{sec:metarules}

In addition to the annotated data,
we also make available all \texttildelow{}960k \resource{}-InfCands
found by our unsupervised algorithm.
\candidates{}'s distribution is  similar enough to our labeled dataset
to be useful for domain adaptation, representation learning
and other techniques when working on LIiC.
It can also be investigated on its own in a purely
unsupervised fashion as we will show now.

We can find easily interpretable patterns by looking for
cases where premise and hypothesis of an InfCand
have common parts.
By masking these
parts  (\texttt{X}), we can abstract away from concrete
instances and interesting meta rules emerge
(\cref{tab:metarules}).  The most common patterns
represent reasonable equivalent formulations, e.g.,
active/passive voice or ``be $Y$'s $X$ $\Leftrightarrow$ be
$X$ of $Y$'' (the \emph{in}-variant coming from a lot of
location-typed rule instances).  The fifth most frequent
pattern could still be formulated in an even more abstract
way but shows already that the general principle of a
conjunction $Y\land X$ implying one of its components $X$
can be learned.

If we search for meta rules whose
\texttt{X} is part of a lemma (rather than a longer  dependency path), we  discover  cases of
derivational morphology such as agent nouns
(e.g., \emph{ruler}, \emph{leader}) and sense preserving verb
prefixes (e.g., \emph{re-write}, \emph{over-react}).

Finally, we observe several implicative verbs (verbs that entail their complement clauses) in their typical pattern $V$ \textit{to} $X \Rightarrow X$.
A lot of these verbs are not traditional implicatives, but are
called
\emph{de facto implicatives} by \citet{pavlick16}
-- who argue for the importance  
of data-driven approaches to
detecting de facto implicatives.
The meta rule discovery method just described is
such a data-driven approach.


\section{Related Work}
\label{sec:related-work}

\minip{NLI challenge datasets.}
A lot of work exists
that aims at uncovering weaknesses in state-of-the-art NLI models.
Several approaches are based on modifications of popular datasets,
such as SNLI or MultiNLI.
These modifications range
from simple rule-based transformations \citep{naik18}
to rewritings generated by genetic algorithms \citep{alzantot18}
or adversarial neural networks \citep{zhao18}.
\citet{lalor16} constructed an NLI test set
by judging the difficulty of the sentence pairs in a small SNLI subset
based on crowd-sourced human responses
via Item Response Theory.
These works are related
as they, too, challenge existing NLI models with new data
but orthogonal to ours
as their goal is not to measure a model's knowledge about
lexical inference in context.

\citet{glockner18} modified SNLI
by replacing one word from a given sentence
by a synonym, (co-)hyponym, hypernym or antonym
to build a test set
that requires NLI systems
to use lexical knowledge.
They rely on WordNet's lexical taxonomy.
This, however, is difficult for verbs
because
their semantics depends  more on  context.
Finally, \citet{glockner18}'s dataset has a strong bias for \emph{contradiction}
whereas our dataset is specifically designed to contain cases of \emph{entailment}.

Our work is more closely related to the dataset by \citet{levy16},
who 
frame relation entailment  as the task of judging the appropriateness of candidate answers.
Their hypothesis is that an answer is only appropriate
if it entails the predicate of the question.
This is often but by no means always true;
certain questions imply additional information.
Consider:
``Which country annexed country[$B$]?''
The answer candidate
``country[$A$] administers country[$B$]''
might be considered valid,
given that it is unlikely
that one country annexes $B$
and another country administers it.
The inference \emph{administer} $\Rightarrow$ \emph{annex},
however, does not hold.
Because of these difficulties,
we follow the more traditional approach \citep{zeichner12}
of asking about consequences of a given fact (the premise).

\minip{Relation extraction.}
Some works \citep{schoenmackers10,berantThesis,zeichner12} rely on the output
from open information extraction systems \citep{banko07,fader11}.
A more flexible approach is to represent relations
as lexicalized paths in dependency graphs \citep{lin01,szpektor04},
sometimes with semantic postprocessing \citep{shwartz17-coref}
and/or retransforming into textual patterns \citep{patty}.
We, too, choose the latter.

\minip{Relation typing.}
Typing relations has become standard in inference mining
because of its usefulness for sense disambiguation
\citep{schoenmackers10,patty,yao12,lewis13}.
Still some resources only provide types
for one argument slot of their binary relations \citep{levy16}
or no types at all \citep{zeichner12,berantThesis,shwartz17-coref}.
Our InfCands are typed in both argument slots,
which both facilitates disambiguation and makes them more general.

Some works \citep{yao12,lewis13} learn
distributions over latent type signatures
for their relations via topic modeling.
A large disadvantage of latent types is
their lack of intuitive interpretability.
By design, our KG types are meaningful and human-interpretable.

\citet{schoenmackers10} type common nouns
based on cooccurrence with class nouns
identified by Hearst patterns \citep{hearst92}
and later try to filter out unreasonable typings
by using frequency thresholds and PMI.
As KG entities are manually labeled with their correct types,
we do not need this kind of heuristics.
Furthermore, in contrast to this ad-hoc type system,
KG types are the result of a KG design process.
Notably, Freebase types function as interfaces, i.e.,
permit type-specific properties to be added,
and are thus inherently motivated by relations between entities.

Lexical ontologies, such as WordNet \citep[as used by][]{levy16}
likewise lack this connection between relations and types.
Moreover,
relations between real-world entities are more often events
than relations between common nouns.
Thus, in contrast to existing resources
that do not restrict
relations to KG entities,
\resource{} contains more event-like relations.

\citet{patty} also use KG types as context for their textual patterns.
They simply create a new relation for each possible type combination
for each entity occurring with a pattern and each possible type of this entity.
It is unclear how the combinatorial explosion and the resulting sparsity
affects pattern quality.
Our approach of successively splitting a typewise heterogenous relation
into its $k$ largest homogenous subrelations
aims at finding only the most typical types for an action
and our definition of type signature as intersection of all common types
avoids unnecessary redundancy.

\minip{Entailment candidate collection.}
Distributional features are a common choice for paraphrase detection
and relation clustering \citep{lin01,szpektor04,sekine05,yao12,lewis13}.

The two most important alternatives
are bilingual pivoting \citep{ppdb-origin}
-- which identifies identically translated phrases in bilingual corpora --
and event coreference in the news \citep{xu14,zhang15,shwartz17-coref}
-- which relies on lexical variability in two articles or headlines
referring to the same event.
We specifically focus on distributional information for our InfCand collection
because current models of lexical semantics are also mainly based on that \citep[e.g.,][]{fasttext}.
Our goal is not
to build a resource free of typical mistakes made by distributional approaches
but to provide a benchmark to study the progress on overcoming them (cf.\ \cref{sec:discussion}).

Another difference to aforementioned works
is that we explicitly model unidirectional entailment
as opposed to bidirectional synonymy (cf.\ \cref{tab:false_positives}, (1)).
Here one can distinguish a learning-based approach \citep{berantThesis},
where an SVM classifier with various features is trained on lexical ontologies like WordNet,
followed by the application of global transitivity constrains to enhance consistency,
and probabilistic models of noisy set inclusion in the tradition of the distributional inclusion hypothesis \citep{schoenmackers10,patty}.
We adapt Sherlock, an instance of the latter,
for its simplicity and effectiveness.


\section{Conclusion}
\label{sec:conclusion}

  We presented \resource{}, a new challenging testbed for LIiC and NLI,
  based on typed textual relations between named entities (NEs) from a KG.
  The restriction to NEs (as opposed to common nouns)
  allowed us to harness more event-like relations than previous similar collections
  as these naturally occur more often with NEs.
  The distributional similarity of both positive and negative examples
  makes \resource{} a promising benchmark to track future NLI models' ability
  to go beyond shallow semantics relying primarily on distributional evidence.
  We showed that existing rule bases are complementary to \resource{}
  and that current semantic vector space models
  as well as SOTA neural NLI models
  cannot achieve at the same time high precision and high recall on \resource{}.
  Although \resource{}'s creation is entirely data-driven,
  it shows a large variety of linguistic challenges for NLI,
  ranging
  from lexical relations like troponymy, synonymy or morph.\ derivation
  to typical actions and common sense knowledge (cf.\ \cref{tab:phenomena}).
  The large unlabeled resources, \candidates{} and \relext{},
  are potentially useful for further linguistic analysis
  (as we showed in \cref{sec:metarules}),
  as well as for data-driven models of lexical semantics, e.g.,
  techniques such as representation learning and domain adaptation.
  We hope that \resource{} will foster better modeling of
  lexical inference  in context as well
  as progress in NLI in general.


\section*{Acknowledgments}
\label{sec:acknowledgments}

We gratefully acknowledge a Ph.D.\ scholarship awarded to the first author
by the German Academic Scholarship Foundation (Studienstiftung des deutschen Volkes).
This work was supported by the BMBF as part of the project MLWin (01IS18050).


\bibliographystyle{acl_natbib}
\bibliography{references}

\begin{thebibliography}{55}
\expandafter\ifx\csname natexlab\endcsname\relax\def\natexlab#1{#1}\fi

\bibitem[{Alzantot et~al.(2018)Alzantot, Sharma, Elgohary, Ho, Srivastava, and
  Chang}]{alzantot18}
Moustafa Alzantot, Yash Sharma, Ahmed Elgohary, Bo-Jhang Ho, Mani Srivastava,
  and Kai-Wei Chang. 2018.
\newblock \href {https://www.aclweb.org/anthology/D18-1316} {Generating natural
  language adversarial examples}.
\newblock In \emph{Proceedings of the 2018 Conference on Empirical Methods in
  Natural Language Processing}, pages 2890--2896, Brussels, Belgium.
  Association for Computational Linguistics.

\bibitem[{Banko et~al.(2007)Banko, Cafarella, Soderland, Broadhead, and
  Etzioni}]{banko07}
Michele Banko, Michael~J Cafarella, Stephen Soderland, Matthew Broadhead, and
  Oren Etzioni. 2007.
\newblock Open information extraction from the web.
\newblock In \emph{Procs. of IJCAI}, volume~7, pages 2670--2676.

\bibitem[{Berant(2012)}]{berantThesis}
Jonathan Berant. 2012.
\newblock \emph{Global Learning of Textual Entailment Graphs}.
\newblock Ph.D. thesis, The Blavatnik School of Computer Science, Raymond and
  Beverly Sackler Faculty of Exact Sciences, Tel Aviv University.

\bibitem[{Berant et~al.(2011)Berant, Dagan, and Goldberger}]{berant11}
Jonathan Berant, Ido Dagan, and Jacob Goldberger. 2011.
\newblock Global learning of typed entailment rules.
\newblock In \emph{Proceedings of the 49th Annual Meeting of the Association
  for Computational Linguistics: Human Language Technologies - Volume 1}, HLT
  '11, pages 610--619, Stroudsburg, PA, USA. Association for Computational
  Linguistics.

\bibitem[{Bollacker et~al.(2008)Bollacker, Evans, Paritosh, Sturge, and
  Taylor}]{freebase}
Kurt Bollacker, Colin Evans, Praveen Paritosh, Tim Sturge, and Jamie Taylor.
  2008.
\newblock Freebase: A collaboratively created graph database for structuring
  human knowledge.
\newblock In \emph{Proceedings of the 2008 ACM SIGMOD International Conference
  on Management of Data}, SIGMOD '08, pages 1247--1250, New York, NY, USA.

\bibitem[{Bordes et~al.(2013)Bordes, Usunier, Garcia-Duran, Weston, and
  Yakhnenko}]{transE}
Antoine Bordes, Nicolas Usunier, Alberto Garcia-Duran, Jason Weston, and Oksana
  Yakhnenko. 2013.
\newblock Translating embeddings for modeling multi-relational data.
\newblock In C.~J.~C. Burges, L.~Bottou, M.~Welling, Z.~Ghahramani, and K.~Q.
  Weinberger, editors, \emph{Advances in Neural Information Processing Systems
  26}, pages 2787--2795. Curran Associates, Inc.

\bibitem[{Bowman et~al.(2015)Bowman, Angeli, Potts, and Manning}]{snli}
Samuel~R. Bowman, Gabor Angeli, Christopher Potts, and Christopher~D. Manning.
  2015.
\newblock A large annotated corpus for learning natural language inference.
\newblock In \emph{Proceedings of the 2015 Conference on Empirical Methods in
  Natural Language Processing (EMNLP)}. Association for Computational
  Linguistics.

\bibitem[{Chen et~al.(2017)Chen, Zhu, Ling, Wei, Jiang, and Inkpen}]{chen17}
Qian Chen, Xiaodan Zhu, Zhen-Hua Ling, Si~Wei, Hui Jiang, and Diana Inkpen.
  2017.
\newblock Enhanced lstm for natural language inference.
\newblock In \emph{Proceedings of the 55th Annual Meeting of the Association
  for Computational Linguistics (Volume 1: Long Papers)}, pages 1657--1668.
  Association for Computational Linguistics.

\bibitem[{Dagan et~al.(2013)Dagan, Roth, Sammons, and Zanzotto}]{dagan13}
Ido Dagan, Dan Roth, Mark Sammons, and Fabio~Massimo Zanzotto. 2013.
\newblock \emph{Recognizing textual entailment: Models and applications}.
\newblock Morgan \& Claypool Publishers.

\bibitem[{Fader et~al.(2011)Fader, Soderland, and Etzioni}]{fader11}
Anthony Fader, Stephen Soderland, and Oren Etzioni. 2011.
\newblock \href {https://www.aclweb.org/anthology/D11-1142} {Identifying
  relations for open information extraction}.
\newblock In \emph{Proceedings of the 2011 Conference on Empirical Methods in
  Natural Language Processing}, pages 1535--1545, Edinburgh, Scotland, UK.
  Association for Computational Linguistics.

\bibitem[{Fellbaum(2005)}]{wordnet05}
Christiane Fellbaum. 2005.
\newblock Wordnet and wordnets.
\newblock In Keith Brown~et al., editor, \emph{Encyclopedia of Language and
  Linguistics}, second edition, pages 665--670. Elsevier, Oxford.

\bibitem[{Fr\'{e}nay and Verleysen(2014)}]{frenay14}
Beno\^{i}t Fr\'{e}nay and Michel Verleysen. 2014.
\newblock {Classification in the Presence of Label Noise: A Survey}.
\newblock \emph{IEEE TRANSACTIONS ON NEURAL NETWORKS AND LEARNING SYSTEMS},
  25(5):845--869.

\bibitem[{Gabrilovich et~al.(2013)Gabrilovich, Ringgaard, and
  Subramanya}]{gabrilovich13}
Evgeniy Gabrilovich, Michael Ringgaard, and Amarnag Subramanya. 2013.
\newblock {FACC1: Freebase annotation of ClueWeb corpora, Version 1 (Release
  date 2013-06-26, Format version 1, Correction level 0)}.

\bibitem[{Ganitkevitch et~al.(2013)Ganitkevitch, Van~Durme, and
  Callison-Burch}]{ppdb-origin}
Juri Ganitkevitch, Benjamin Van~Durme, and Chris Callison-Burch. 2013.
\newblock \href {https://www.aclweb.org/anthology/N13-1092} {{PPDB}: The
  paraphrase database}.
\newblock In \emph{Proceedings of the 2013 Conference of the North {A}merican
  Chapter of the Association for Computational Linguistics: Human Language
  Technologies}, pages 758--764, Atlanta, Georgia. Association for
  Computational Linguistics.

\bibitem[{Glockner et~al.(2018)Glockner, Shwartz, and Goldberg}]{glockner18}
Max Glockner, Vered Shwartz, and Yoav Goldberg. 2018.
\newblock Breaking {NLI} systems with sentences that require simple lexical
  inferences.
\newblock In \emph{The 56th Annual Meeting of the Association for Computational
  Linguistics (ACL)}, Melbourne, Australia.

\bibitem[{Grave et~al.(2017)Grave, Mikolov, Joulin, and Bojanowski}]{fasttext}
Edouard Grave, Tomas Mikolov, Armand Joulin, and Piotr Bojanowski. 2017.
\newblock Bag of tricks for efficient text classification.
\newblock In \emph{{EACL} {(2)}}, pages 427--431. Association for Computational
  Linguistics.

\bibitem[{Gururangan et~al.(2018)Gururangan, Swayamdipta, Levy, Schwartz,
  Bowman, and Smith}]{gururangan18}
Suchin Gururangan, Swabha Swayamdipta, Omer Levy, Roy Schwartz, Samuel Bowman,
  and Noah~A. Smith. 2018.
\newblock \href {http://aclweb.org/anthology/N18-2017} {Annotation artifacts in
  natural language inference data}.
\newblock In \emph{Proceedings of the 2018 Conference of the North American
  Chapter of the Association for Computational Linguistics: Human Language
  Technologies, Volume 2 (Short Papers)}, pages 107--112. Association for
  Computational Linguistics.

\bibitem[{Han et~al.(2018)Han, Cao, Xin, Lin, Liu, Sun, and Li}]{openKE}
Xu~Han, Shulin Cao, Lv~Xin, Yankai Lin, Zhiyuan Liu, Maosong Sun, and Juanzi
  Li. 2018.
\newblock Openke: An open toolkit for knowledge embedding.
\newblock In \emph{Proceedings of EMNLP}.

\bibitem[{Hearst(1992)}]{hearst92}
Marti~A. Hearst. 1992.
\newblock Automatic acquisition of hyponyms from large text corpora.
\newblock In \emph{Proceedings of the 14th Conference on Computational
  Linguistics - Volume 2}, COLING '92, pages 539--545, Stroudsburg, PA, USA.
  Association for Computational Linguistics.

\bibitem[{Hendrycks et~al.(2018)Hendrycks, Mazeika, Wilson, and
  Gimpel}]{hendrycks18}
Dan Hendrycks, Mantas Mazeika, Duncan Wilson, and Kevin Gimpel. 2018.
\newblock Using trusted data to train deep networks on labels corrupted by
  severe noise.
\newblock In S.~Bengio, H.~Wallach, H.~Larochelle, K.~Grauman, N.~Cesa-Bianchi,
  and R.~Garnett, editors, \emph{Advances in Neural Information Processing
  Systems 31}, pages 10477--10486. Curran Associates, Inc.

\bibitem[{Kotlerman et~al.(2010)Kotlerman, Dagan, Szpektor, and
  Zhitomirsky-geffet}]{kotlerman10}
Lili Kotlerman, Ido Dagan, Idan Szpektor, and Maayan Zhitomirsky-geffet. 2010.
\newblock Directional distributional similarity for lexical inference.
\newblock \emph{Nat. Lang. Eng.}, 16(4):359--389.

\bibitem[{Lalor et~al.(2016)Lalor, Wu, and yu}]{lalor16}
John Lalor, Hao Wu, and hong yu. 2016.
\newblock \href {https://aclweb.org/anthology/D16-1062} {Building an evaluation
  scale using item response theory}.
\newblock In \emph{Proceedings of the 2016 Conference on Empirical Methods in
  Natural Language Processing}, pages 648--657, Austin, Texas. Association for
  Computational Linguistics.

\bibitem[{Lenci and Benotto(2012)}]{lenci12}
Alessandro Lenci and Giulia Benotto. 2012.
\newblock Identifying hypernyms in distributional semantic spaces.
\newblock In \emph{{*SEM 2012}: The First Joint Conference on Lexical and
  Computational Semantics -- Volume 1: Proceedings of the main conference and
  the shared task, and Volume 2: Proceedings of the Sixth International
  Workshop on Semantic Evaluation {(SemEval 2012)}}, pages 75--79,
  Montr\'{e}al, Canada. Association for Computational Linguistics.

\bibitem[{Levy and Dagan(2016)}]{levy16}
Omer Levy and Ido Dagan. 2016.
\newblock Annotating relation inference in context via question answering.
\newblock In \emph{Proceedings of the 54th Annual Meeting of the Association
  for Computational Linguistics, {ACL} 2016, August 7-12, 2016, Berlin,
  Germany, Volume 2: Short Papers}.

\bibitem[{Lewis and Steedman(2013)}]{lewis13}
Mike Lewis and Mark Steedman. 2013.
\newblock Combining distributional and logical semantics.
\newblock \emph{Transactions of the Association for Computational Linguistics},
  1:179--192.

\bibitem[{Lin and Pantel(2001)}]{lin01}
Dekang Lin and Patrick Pantel. 2001.
\newblock \href {http://www.cs.ualberta.ca/~lindek/papers/kdd01-1.pdf} {{DIRT}:
  {D}iscovery of {I}nference {R}ules from {T}ext}.
\newblock In \emph{Proceedings of the Seventh ACM SIGKDD International
  Conference on Knowledge Discovery and Data Mining (KDD'01)}, pages 323--328,
  New York, NY, USA. ACM Press.

\bibitem[{Loper and Bird(2002)}]{nltk}
Edward Loper and Steven Bird. 2002.
\newblock Nltk: The natural language toolkit.
\newblock In \emph{In Proceedings of the ACL Workshop on Effective Tools and
  Methodologies for Teaching Natural Language Processing and Computational
  Linguistics. Philadelphia: Association for Computational Linguistics}.

\bibitem[{Mikolov et~al.(2013{\natexlab{a}})Mikolov, Chen, Corrado, and
  Dean}]{word2vec_alg}
Tomas Mikolov, Kai Chen, Greg Corrado, and Jeffrey Dean. 2013{\natexlab{a}}.
\newblock \href {http://arxiv.org/abs/1301.3781} {Efficient estimation of word
  representations in vector space}.
\newblock \emph{CoRR}, abs/1301.3781.

\bibitem[{Mikolov et~al.(2013{\natexlab{b}})Mikolov, Sutskever, Chen, Corrado,
  and Dean}]{word2vec_emb}
Tomas Mikolov, Ilya Sutskever, Kai Chen, Greg~S Corrado, and Jeff Dean.
  2013{\natexlab{b}}.
\newblock \href
  {http://papers.nips.cc/paper/5021-distributed-representations-of-words-and-phrases-and-their-compositionality.pdf}
  {Distributed representations of words and phrases and their
  compositionality}.
\newblock In C.~J.~C. Burges, L.~Bottou, M.~Welling, Z.~Ghahramani, and K.~Q.
  Weinberger, editors, \emph{Advances in Neural Information Processing Systems
  26}, pages 3111--3119. Curran Associates, Inc.

\bibitem[{Miller(1995)}]{wordnet95}
George~A. Miller. 1995.
\newblock Wordnet: A lexical database for english.
\newblock \emph{Commun. ACM}, 38(11):39--41.

\bibitem[{Mohammad et~al.(2008)Mohammad, Dorr, and Hirst}]{mohammad08}
Saif Mohammad, Bonnie Dorr, and Graeme Hirst. 2008.
\newblock \href {http://www.aclweb.org/anthology/D08-1103} {Computing word-pair
  antonymy}.
\newblock In \emph{Proceedings of the 2008 Conference on Empirical Methods in
  Natural Language Processing}, pages 982--991, Honolulu, Hawaii. Association
  for Computational Linguistics.

\bibitem[{Mrk\v{s}i\'{c} et~al.(2016)Mrk\v{s}i\'{c}, \'{O}~S\'{e}aghdha,
  Thomson, Ga\v{s}i\'{c}, Rojas-Barahona, Su, Vandyke, Wen, and
  Young}]{mrkvsic16}
Nikola Mrk\v{s}i\'{c}, Diarmuid \'{O}~S\'{e}aghdha, Blaise Thomson, Milica
  Ga\v{s}i\'{c}, Lina~M. Rojas-Barahona, Pei-Hao Su, David Vandyke, Tsung-Hsien
  Wen, and Steve Young. 2016.
\newblock \href {http://www.aclweb.org/anthology/N16-1018} {Counter-fitting
  word vectors to linguistic constraints}.
\newblock In \emph{Proceedings of the 2016 Conference of the North American
  Chapter of the Association for Computational Linguistics: Human Language
  Technologies}, pages 142--148, San Diego, California. Association for
  Computational Linguistics.

\bibitem[{Naik et~al.(2018)Naik, Ravichander, Sadeh, Rose, and Neubig}]{naik18}
Aakanksha Naik, Abhilasha Ravichander, Norman Sadeh, Carolyn Rose, and Graham
  Neubig. 2018.
\newblock \href {https://www.aclweb.org/anthology/C18-1198} {Stress test
  evaluation for natural language inference}.
\newblock In \emph{Proceedings of the 27th International Conference on
  Computational Linguistics}, pages 2340--2353, Santa Fe, New Mexico, USA.
  Association for Computational Linguistics.

\bibitem[{Nakashole et~al.(2012)Nakashole, Weikum, and Suchanek}]{patty}
Ndapandula Nakashole, Gerhard Weikum, and Fabian Suchanek. 2012.
\newblock \href {http://www.aclweb.org/anthology/D12-1104} {Patty: A taxonomy
  of relational patterns with semantic types}.
\newblock In \emph{Proceedings of the 2012 Joint Conference on Empirical
  Methods in Natural Language Processing and Computational Natural Language
  Learning}, pages 1135--1145, Jeju Island, Korea. Association for
  Computational Linguistics.

\bibitem[{Nivre et~al.(2007)Nivre, Hall, Nilsson, Chanev, Eryigit, Kübler,
  Marinov, and Marsi}]{malt07}
Joakim Nivre, Johan Hall, Jens Nilsson, Atanas Chanev, Gülşen Eryigit, Sandra
  Kübler, Svetoslav Marinov, and Erwin Marsi. 2007.
\newblock \href {https://doi.org/10.1017/S1351324906004505} {Maltparser: A
  language-independent system for data-driven dependency parsing}.
\newblock \emph{Natural Language Engineering}, 13(2):95–135.

\bibitem[{Pavlick and Callison-Burch(2016)}]{pavlick16}
Ellie Pavlick and Chris Callison-Burch. 2016.
\newblock \href {https://aclweb.org/anthology/D16-1240} {Tense manages to
  predict implicative behavior in verbs}.
\newblock In \emph{Proceedings of the 2016 Conference on Empirical Methods in
  Natural Language Processing}, pages 2225--2229, Austin, Texas. Association
  for Computational Linguistics.

\bibitem[{Pavlick et~al.(2015)Pavlick, Rastogi, Ganitkevitch, Van~Durme, and
  Callison-Burch}]{ppdb}
Ellie Pavlick, Pushpendre Rastogi, Juri Ganitkevitch, Benjamin Van~Durme, and
  Chris Callison-Burch. 2015.
\newblock \href {http://www.aclweb.org/anthology/P15-2070} {Ppdb 2.0: Better
  paraphrase ranking, fine-grained entailment relations, word embeddings, and
  style classification}.
\newblock In \emph{Proceedings of the 53rd Annual Meeting of the Association
  for Computational Linguistics and the 7th International Joint Conference on
  Natural Language Processing (Volume 2: Short Papers)}, pages 425--430,
  Beijing, China. Association for Computational Linguistics.

\bibitem[{Roller et~al.(2018)Roller, Kiela, and Nickel}]{roller18}
Stephen Roller, Douwe Kiela, and Maximilian Nickel. 2018.
\newblock Hearst patterns revisited: Automatic hypernym detection from large
  text corpora.
\newblock In \emph{Proceedings of the 56th Annual Meeting of the Association
  for Computational Linguistics (Volume 2: Short Papers)}, pages 358--363,
  Melbourne, Australia. Association for Computational Linguistics.

\bibitem[{Salmon(1971)}]{salmon71}
Wesley~C Salmon. 1971.
\newblock \emph{Statistical explanation and statistical relevance}, volume~69.
\newblock University of Pittsburgh Pre.

\bibitem[{Santus et~al.(2014)Santus, Lenci, Lu, and Schulte~im
  Walde}]{santus14}
Enrico Santus, Alessandro Lenci, Qin Lu, and Sabine Schulte~im Walde. 2014.
\newblock \href {http://www.aclweb.org/anthology/E14-4008} {Chasing hypernyms
  in vector spaces with entropy}.
\newblock In \emph{Proceedings of the 14th Conference of the European Chapter
  of the Association for Computational Linguistics, volume 2: Short Papers},
  pages 38--42, Gothenburg, Sweden. Association for Computational Linguistics.

\bibitem[{Schoenmackers et~al.(2010)Schoenmackers, Davis, Etzioni, and
  Weld}]{schoenmackers10}
Stefan Schoenmackers, Jesse Davis, Oren Etzioni, and Daniel Weld. 2010.
\newblock Learning first-order horn clauses from web text.
\newblock In \emph{Proceedings of the Conference on Empirical Methods in
  Natural Language Processing: 2010}, pages 1088--1098.

\bibitem[{Sekine(2005)}]{sekine05}
Satoshi Sekine. 2005.
\newblock \href {https://www.aclweb.org/anthology/I05-5011} {Automatic
  paraphrase discovery based on context and keywords between {NE} pairs}.
\newblock In \emph{Proceedings of the Third International Workshop on
  Paraphrasing ({IWP}2005)}.

\bibitem[{Shwartz et~al.(2017{\natexlab{a}})Shwartz, Santus, and
  Schlechtweg}]{shwartz17}
Vered Shwartz, Enrico Santus, and Dominik Schlechtweg. 2017{\natexlab{a}}.
\newblock Hypernyms under siege: Linguistically-motivated artillery for
  hypernymy detection.
\newblock In \emph{Proceedings of the 15th Conference of the European Chapter
  of the Association for Computational Linguistics: Volume 1, Long Papers},
  pages 65--75, Valencia, Spain. Association for Computational Linguistics.

\bibitem[{Shwartz et~al.(2017{\natexlab{b}})Shwartz, Stanovsky, and
  Dagan}]{shwartz17-coref}
Vered Shwartz, Gabriel Stanovsky, and Ido Dagan. 2017{\natexlab{b}}.
\newblock \href {https://www.aclweb.org/anthology/S17-1019} {Acquiring
  predicate paraphrases from news tweets}.
\newblock In \emph{Proceedings of the 6th Joint Conference on Lexical and
  Computational Semantics (*{SEM} 2017)}, pages 155--160, Vancouver, Canada.
  Association for Computational Linguistics.

\bibitem[{Szpektor et~al.(2004)Szpektor, Tanev, Dagan, and
  Coppola}]{szpektor04}
Idan Szpektor, Hristo Tanev, Ido Dagan, and Bonaventura Coppola. 2004.
\newblock \href {https://www.aclweb.org/anthology/W04-3206} {Scaling web-based
  acquisition of entailment relations}.
\newblock In \emph{Proceedings of {EMNLP} 2004}, pages 41--48, Barcelona,
  Spain. Association for Computational Linguistics.

\bibitem[{Trouillon et~al.(2016)Trouillon, Welbl, Riedel, Gaussier, and
  Bouchard}]{complex}
Th{\'e}o Trouillon, Johannes Welbl, Sebastian Riedel, \'{E}ric Gaussier, and
  Guillaume Bouchard. 2016.
\newblock Complex embeddings for simple link prediction.
\newblock In \emph{Proceedings of the 33rd International Conference on
  International Conference on Machine Learning - Volume 48}, ICML'16, pages
  2071--2080.

\bibitem[{Weeds et~al.(2004)Weeds, Weir, and McCarthy}]{weeds04}
Julie Weeds, David Weir, and Diana McCarthy. 2004.
\newblock Characterising measures of lexical distributional similarity.
\newblock In \emph{Proceedings of Coling 2004}, pages 1015--1021, Geneva,
  Switzerland. COLING.

\bibitem[{Weissenborn et~al.(2018)Weissenborn, Minervini, Dettmers, Augenstein,
  Welbl, Rocktäschel, Bošnjak, Mitchell, Demeester, Stenetorp, and
  Riedel}]{weissenborn18}
Dirk Weissenborn, Pasquale Minervini, Tim Dettmers, Isabelle Augenstein,
  Johannes Welbl, Tim Rocktäschel, Matko Bošnjak, Jeff Mitchell, Thomas
  Demeester, Pontus Stenetorp, and Sebastian Riedel. 2018.
\newblock \href {https://arxiv.org/abs/1806.08727} {{Jack the Reader – A
  Machine Reading Framework}}.
\newblock In \emph{{Proceedings of the 56th Annual Meeting of the Association
  for Computational Linguistics (ACL) System Demonstrations}}.

\bibitem[{Williams et~al.(2018)Williams, Nangia, and Bowman}]{multinli}
Adina Williams, Nikita Nangia, and Samuel Bowman. 2018.
\newblock \href {http://aclweb.org/anthology/N18-1101} {A broad-coverage
  challenge corpus for sentence understanding through inference}.
\newblock In \emph{Proceedings of the 2018 Conference of the North American
  Chapter of the Association for Computational Linguistics: Human Language
  Technologies, Volume 1 (Long Papers)}, pages 1112--1122. Association for
  Computational Linguistics.

\bibitem[{{XTAG Research Group}(2001)}]{xtag01}
{XTAG Research Group}. 2001.
\newblock A lexicalized tree adjoining grammar for english.
\newblock Technical Report IRCS-01-03, IRCS, University of Pennsylvania.

\bibitem[{Xu et~al.(2014)Xu, Ritter, Callison-Burch, Dolan, and Ji}]{xu14}
Wei Xu, Alan Ritter, Chris Callison-Burch, William~B. Dolan, and Yangfeng Ji.
  2014.
\newblock \href {https://www.aclweb.org/anthology/Q14-1034} {Extracting
  lexically divergent paraphrases from twitter}.
\newblock \emph{Transactions of the Association for Computational Linguistics},
  2:435--448.

\bibitem[{Yao et~al.(2012)Yao, Riedel, and McCallum}]{yao12}
Limin Yao, Sebastian Riedel, and Andrew McCallum. 2012.
\newblock Unsupervised relation discovery with sense disambiguation.
\newblock In \emph{Proceedings of the 50th Annual Meeting of the Association
  for Computational Linguistics: Long Papers - Volume 1}, ACL '12, pages
  712--720, Stroudsburg, PA, USA. Association for Computational Linguistics.

\bibitem[{Zeichner et~al.(2012)Zeichner, Berant, and Dagan}]{zeichner12}
Naomi Zeichner, Jonathan Berant, and Ido Dagan. 2012.
\newblock \href {https://www.aclweb.org/anthology/P12-2031} {Crowdsourcing
  inference-rule evaluation}.
\newblock In \emph{Proceedings of the 50th Annual Meeting of the Association
  for Computational Linguistics (Volume 2: Short Papers)}, pages 156--160, Jeju
  Island, Korea. Association for Computational Linguistics.

\bibitem[{Zhang et~al.(2015)Zhang, Soderland, and Weld}]{zhang15}
Congle Zhang, Stephen Soderland, and Daniel~S. Weld. 2015.
\newblock \href {https://www.aclweb.org/anthology/Q15-1009} {Exploiting
  parallel news streams for unsupervised event extraction}.
\newblock \emph{Transactions of the Association for Computational Linguistics},
  3:117--129.

\bibitem[{Zhao et~al.(2018)Zhao, Dua, and Singh}]{zhao18}
Zhengli Zhao, Dheeru Dua, and Sameer Singh. 2018.
\newblock \href {https://openreview.net/forum?id=H1BLjgZCb} {Generating natural
  adversarial examples}.
\newblock In \emph{International Conference on Learning Representations}.

\end{thebibliography}

\appendix

\section{Relation Filter Heuristics}
\label{app:relation-filter}
In order to be kept as a relation, a dependency path must fulfill all of the following criteria:
\begin{enumerate}
\item It starts or ends with \texttt{nsubj} or \texttt{nsubjpass}.
\item It starts or ends with one of the following labels:
  \texttt{nsubj}, \texttt{nsubjpass}, \texttt{iobj}, \texttt{dobj}, \texttt{pobj}, \texttt{appos}, \texttt{poss}, \texttt{rcmod}, \texttt{infmod}, \texttt{partmod}.
\item It is not longer than 7 words and 8 dependency labels.
\item At least one of the presumable lemmas contains at least 3 letters.
\item It does not have the same dependency label at both ends.
\item It does not contain any of the following labels:
  \texttt{parataxis}, \texttt{pcomp}, \texttt{csubj}, \texttt{advcl}, \texttt{ccomp}.
\item It does not contain immediate repetitions of words or dependency labels.
\end{enumerate}

\end{document}